\documentclass[10pt,twocolumn,letterpaper]{article}

\usepackage{iccv}              % To produce the CAMERA-READY version
\definecolor{iccvblue}{rgb}{0.21,0.49,0.74}
\usepackage[pagebackref,breaklinks,colorlinks,allcolors=iccvblue]{hyperref}
\usepackage{multirow} % Allows multirow cells

\usepackage{array}
\usepackage{booktabs}

\def\paperID{SEA-12} % *** Enter the Paper ID here
\def\confName{ICCV}
\def\confYear{2025}

\title{PyroFocus: A Deep Learning Approach to Real-Time Wildfire Detection in Multispectral Remote Sensing Imagery}

\author{
Mark Moussa, Andre Williams, Seth Roffe, Douglas Morton\\
NASA Goddard Space Flight Center\\
{\tt\small \{mark.m.moussa, andreali.williams, seth.roffe, douglas.morton\}@nasa.gov}
}

\begin{document}
\maketitle
\begin{abstract}
% The ABSTRACT is to be in fully justified italicized text, at the top of the left-hand column, below the author and affiliation information.
% Use the word ``Abstract'' as the title, in 12-point Times, boldface type, centered relative to the column, initially capitalized.
% The abstract is to be in 10-point, single-spaced type.
% Leave two blank lines after the Abstract, then begin the main text.
% Look at previous \confName abstracts to get a feel for style and length.

Rapid and accurate wildfire detection and quantification are crucial for emergency response and environmental management. In airborne and spaceborne missions, it is useful for real-time detection algorithms to not only classify whether the observed area consists of no fire, active fire or post-fire, but also to discriminate among varying fire intensities. Hyperspectral and multispectral thermal imagers provide sufficient spectral information to address these challenges; however, high data dimensionality and limited onboard resources complicate real-time processing. With wildfires becoming more common, the need for real-time detection is crucial for successful wildfire management and damage mitigation. It is therefore imperative for onboard solutions to achieve both high accuracy and low inference latency at minimal computational cost. 

In this work, we present several contributions divided into two main parts. First, we systematically evaluate multiple cutting-edge deep learning architectures, such as custom Convolutional Neural Networks (CNN) and Transformer-based models for multi-class fire classification. Second, we propose a novel adaptation of a two-stage pipeline, which we call \textbf{PyroFocus}, consisting of classification of fires and fire radiative power (FRP) regression or image segmentation, designed to significantly reduce inference times and computational requirements for onboard wildfire detection in airborne and spaceborne applications. We evaluate model accuracy, inference latency, and computational cost to identify the most effective solution for resource-constrained environments. Our study leverages a dataset from NASA’s MODIS/ASTER Airborne Simulator (MASTER), which provides data similar to that of an in-house next-generation fire detection sensor, NASA's Compact Fire Imager (CFI) \cite{NASA_CFI}.

Experimental results indicate that the proposed two-stage pipeline provides favorable trade-offs among inference speed and accuracy, demonstrating strong potential for edge deployment in future wildfire monitoring missions.

\end{abstract}    
\section{Introduction}
\label{sec:intro}

% Please follow the steps outlined below when submitting your manuscript to the IEEE Computer Society Press.
% This style guide now has several important modifications (for example, you are no longer warned against the use of sticky tape to attach your artwork to the paper), so all authors should read this new version.

Wildfires rank among the most destructive and rapidly evolving natural disasters, posing severe threats to ecosystems, human lives, and infrastructure. Early detection and precise quantification of wildfire intensity can critically enhance emergency response efforts and resource allocation. Airborne and spaceborne sensor systems, which provide broad geographic coverage at various revisit frequencies, have emerged as vital tools in monitoring wildfires. However, the operational limitations of space platforms, such as slow downlink speeds and resource-constrained computing systems, still act as a barrier for real-time data processing and detection. These barriers can be overcome with the advent of onboard data processing, where algorithms onboard the spacecraft provide the majority of any analysis processing prior to downlink.
%However, operational constraints, particularly when running onboard these platforms, present an array of technical challenges, from real-time data processing to limited computational resources and bandwidth restrictions.

% Forests alone cover about 29\% of the Earth \cite{Seydi22}. Before the advent of remote sensing technologies, wildfire detection generally relied on ground-based lookout towers and human reports leading to delayed response efforts. 

Recent advances in artificial intelligence and machine learning (AI/ML), and in particular, deep learning, have shown promise for extracting complex spectral–spatial features from hyperspectral and multispectral data \cite{yang2018hyperspectral}. However, the high dimensionality and data sizes of these imagery and the necessity for low-latency inference highlight the need for specialized architectures and model pipelines. Furthermore, these deep learning architectures tend to use significant computational resources, which may not be problematic in most cases, but can prove troublesome when dealing with deployment on a heavily resource-constrained system. Balancing accuracy with computational efficiency remains an open challenge.

One of the key methods for characterizing wildfires is Fire Radiative Power (FRP), a well-established metric linked to energy release rates. Accurate FRP estimation can provide further insight into the wildfire dynamics beyond metrics linked to the location and duration of fire activity, such as fire intensity and radiant heat. FRP offers valuable insights into fire behavior, spread potential, and severity. High FRP values indicate intense fires likely to grow rapidly \cite{Laurent}, and thus quantifying FRP can support resource allocation for suppression efforts. Additionally, FRP plays a key role in estimating biomass combustion rates and associated carbon emissions \cite{Li}, supporting studies on air quality and climate impact. In order to provide responsive and precise wildfire management and damage mitigation, these predictive metrics become critical. Therefore, real-time onboard sensor processing needs to keep up with desired response time by providing these metrics with careful attention to inference speed and model efficiency.

In this paper, we tackle these inference latency challenges in two ways: first, we perform a comprehensive evaluation of CNN and Transformer models for multi-class wildfire classification. For this, we elect to focus mainly on model accuracy, since accuracy is ultimately the most important metric when it comes to the detection of wildfires; false positives or false negatives can both have penalizing effects in downstream wildfire mitigation operations. We then utilize the best classification model and compare alongside two distinct strategies for FRP regression: a traditional approach, where we perform regression over every pixel, regardless of whether the patch contains a fire or not, versus PyroFocus, our two-stage pipeline that first classifies fire pixels and then estimates their radiative power. By separating classification from regression, the PyroFocus approach has the potential to significantly reduce the workload, since only the fire-labeled regions require the more computationally expensive FRP analysis. We conduct experiments using data from NASA’s MODIS/ASTER Airborne Simulator (MASTER) \cite{hook2021}, selecting nine specific bands to mirror the capabilities of an in-house sensor under development, NASA's Compact Fire Imager (CFI). We benchmark the approaches with respect to accuracy, inference speed, and computational cost, shedding light on the trade-offs inherent in different architectures and pipeline designs.
\section{Related Works}
\label{sec:relatedworks}

Deep learning has become a powerful tool in remote sensing, advancing tasks such as classification, segmentation, and regression on multispectral and hyperspectral data \cite{bazi2021vision, gretok2021onboard}. Convolutional Neural Networks have proven effective for extracting complex spectral–spatial features, while Transformer-based models leverage attention mechanisms for potentially improved accuracy. However, these deep models often demand significant computational resources, posing challenges for onboard or real-time processing in resource-constrained environments.

In wildfire detection specifically, specialized frameworks such as Fire-Net show that integrating optical and thermal data can achieve high accuracy, even for small fires \cite{Seydi22}. Similar AI-driven systems have also demonstrated rapid alert generation capabilities in severe events like the Australian wildfires \cite{rs15030720}. Comprehensive reviews highlight the versatility of ML methods for broader wildfire science, covering tasks such as fuel characterization, mapping, weather analysis, and fire behavior prediction \cite{wildfirereviewpaper}. Yet these studies also underscore key limitations: balancing accuracy with inference speed, handling class imbalance, and confronting hardware constraints remain open challenges. Recent advancements in early wildfire detection from hyperspectral satellite images \cite{Toan2019} and reviews of deep learning methods for forest fire surveillance systems \cite{Saleh2023} further emphasize these challenges. This work builds on these insights by proposing a two-stage pipeline that emphasizes both accuracy and resource efficiency for real-time wildfire detection and FRP estimation.
\section{Methods}
\label{sec:methods}

This section describes the processes used to achieve two main objectives: evaluating several model performances for the multi-class classification of fire images, and evaluating the efficacy of our proposed two-stage model pipeline (PyroFocus) vs. a traditional one-stage model pipeline.

\subsection{Dataset}
\label{sec:dataset}

\subsubsection{Data source}
\label{sec:data_source}
We utilized data coming from the MASTER 2019 campaign. Specifically, we used the data collected from the Fire Influence on Regional to Global Environments and Air Quality (FIREX-AQ) program, which consists of 21 flights spanning from July to September 2019. We used this dataset specifically because some of the bands provided are similar in nature to CFI, a sensor currently being developed in-house at NASA Goddard Space Flight Center \cite{NASA_CFI}. The MASTER sensor collects multispectral imagery, consisting of 50 bands, covering wavelengths of 0.460 to 12.879 micrometers at varying spatial resolution. The spectral information collected is similar to that provided by the Moderate Resolution Imaging Spectroradiometer (MODIS) onboard Terra and the Advanced Spaceborne Thermal Emission and Reflection Radiometer (ASTER) onboard Terra and Aqua \cite{hook2021}.

% The spectral information collected is also similar to that provided by the Moderate Resolution Imaging Spectroradiometer (MODIS) and the Advanced Spaceborne Thermal Emission and Reflection Radiometer (ASTER), which are aboard two NASA Earth Observing System Satellites, TERRA \& AQUA \cite{hook2021}.

\subsubsection{Band selection}
\label{sec:band_selection}

For our use case, we selected nine of the 50 spectral bands available from the MASTER dataset. The motivation of selecting these bands in particular was not only to closely resemble the data for NASA's CFI, but also to pick the bands best-suited for wildfire detection. We selected bands from the shortwave infrared (SWIR) wavelength, the mid-wave infrared (MWIR) and the long-wave infrared (LWIR) regions. SWIR bands are primarily used here for detecting reflected solar radiation, but these wavelengths are also useful for identifying emitted radiation from hot fires. MWIR and LWIR regions constitute the thermal infrared (TIR) range. MWIR is commonly used in fire detection because it captures the peak emission from high-temperature sources (e.g., flaming fires), while LWIR is useful for capturing smoldering fires, and for background temperature differentiation. Thus, these selected bands are well-suited to detect various aspects of wildfires. Table \ref{tab:spectral_bands} shows the specific bands selected.

\begin{table}[ht]
    \centering
    \begin{tabular}{l c}
        \toprule
        \textbf{Wavelength Region} & \textbf{Spectral Bands ($\mu$m)} \\
        \midrule
        Shortwave Infrared (SWIR) & 2.160, 2.210, 2.260 \\
        Midwave Infrared (MWIR)   & 3.755, 3.910 \\
        Longwave Infrared (LWIR)  & 8.200, 11.330, 12.130 \\
        \bottomrule
    \end{tabular}
    \caption{Grouping of selected spectral bands from the MASTER sensor into wavelength regions.}
    \label{tab:spectral_bands}
\end{table}

\subsubsection{Preprocessing}
\label{sec:preprocessing}

Our data pipeline processes MASTER sensor data stored in Hierarchical Data Format (HDF) files to create analysis-ready datasets. The workflow begins by selecting a subset of HDF files from our multispectral imagery collection. Then, in our preprocessing pipeline, we begin by extracting calibrated spectral measurements and applying radiometric calibration using instrument-specific scale factors. Next, we extract precise geographic coordinates for each pixel to enable accurate geolocation. The resulting 3D data array (height × width × bands) is then converted into a GeoPandas format with wavelength-based column names, ensuring that the data is spatially co-located and well-organized so that the ML training process can effectively leverage spatial features. For FRP estimation, we associate each FRP value with its corresponding pixel by employing a nearest-neighbor spatial join with a 5-meter threshold. Finally, the data is converted back from GeoPandas to NumPy matrices to facilitate faster data-loading performance. This process is only done once, when initially preprocessing the dataset. The resulting dataset is saved out to disk, which will then be used for training, validation, and testing.

The next step we took was implementing a patch-based processing strategy. Each patch has dimensions $H \times W \times C$, where $H$ and $W$ represent the height and width of the patch in pixels, and $C$ is the number of spectral bands used (spectral bands can be found in Table~\ref{tab:spectral_bands}). In our case, we set $H = 24$, $W = 64$, and $C = 9$. Using image patches for training allows for efficient representation of spatially rare events such as fires, facilitates balanced sampling between fire and non-fire classes, and improves overall data coverage. Additionally, patches enhance training via improved sampling of rare events and help maintain a balanced dataset, collectively leading to improved model performance.

% \begin{figure}
%     \centering
%     \includegraphics[width=1\linewidth]{figures/dataaugmentation.pdf}
%     \caption{An example patch and the various ways we augmented. Starting from top left and going sequentially, we have the original patch, then one with Gaussian noise added to it, then one horizontally flipped , then one vertically flipped.}
%     \label{fig:patch-augmentation}
% \end{figure}

In order to prepare the dataset for training and testing, a number of steps were taken. As with most ML data pipelines, data was split into training (80\%), validation (10\%), and testing (10\%) splits. A scaler was fit on the training data only, and used on the training, validation, and test set. This ensures consistent data transformations across all splits, prevents data leakage, and improves model stability by scaling input features \cite{data_scaling_impact}. Three scaling methods—MinMax, Standard, and Robust—were tested. MinMax was chosen as it yielded slightly better performance.

% Due to the high imbalance of no fire, to fire and smoldering pixels and images, dataset augmentation was utilized for the training set. This consisted of flipping horizontally or vertically, and adding Gaussian noise, only to patches that were known to contain fires, in order to bring the training dataset up to a 50/50 split, between no fire, and fire, smoldering, and saturated. Figure \ref{fig:patch-augmentation} shows an example of a patch, and the various ways we augmented. By patching the images beforehand, we ensured a more balanced 50/50 split by selectively augmenting only regions containing fire or smoldering pixels. This approach reduced redundant augmentations of non-fire areas. Figure \ref{fig:class_distribution} shows the distribution of classes before augmentation.

\begin{figure}
    \centering
    \includegraphics[width=0.9\linewidth]{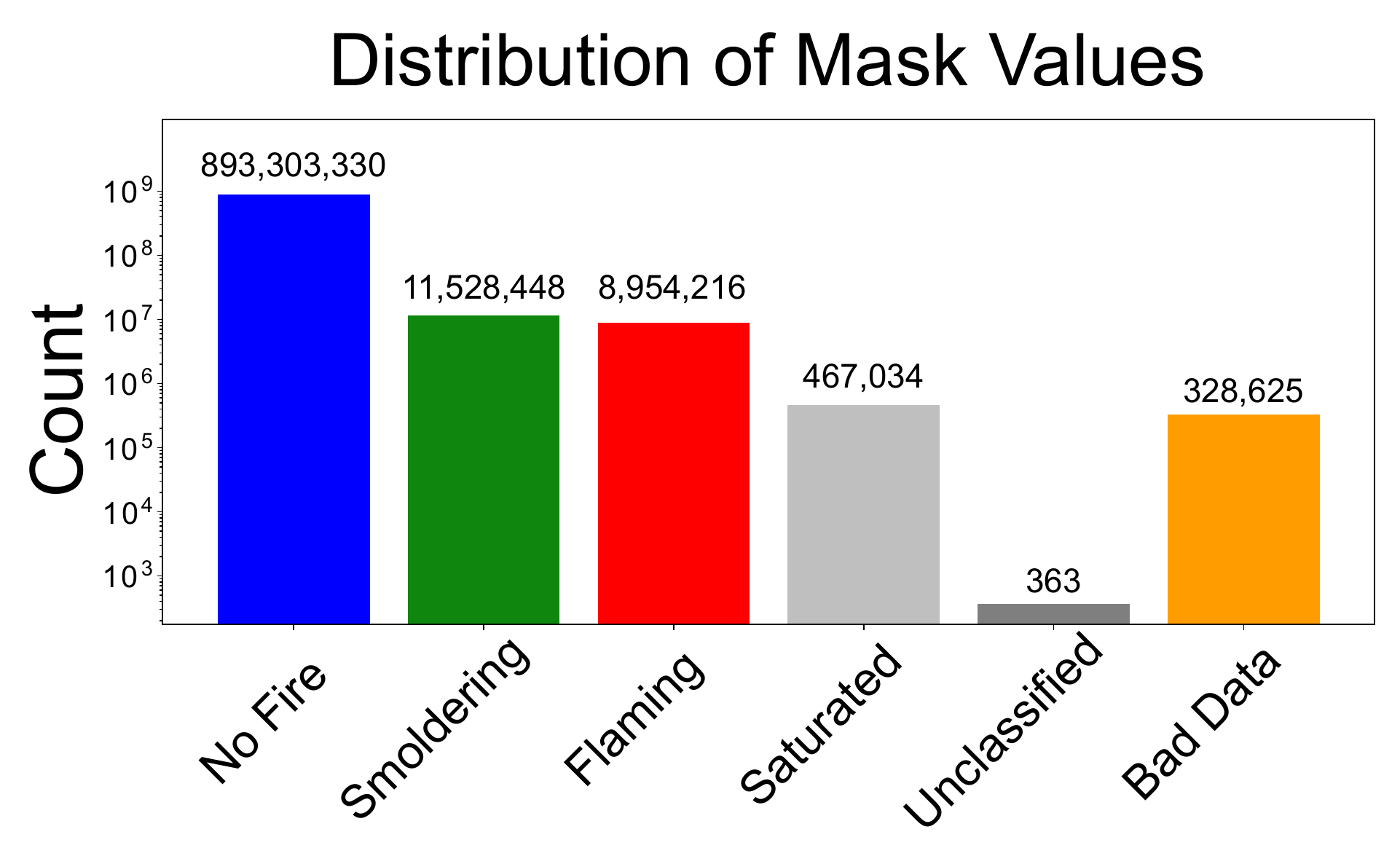}
    \caption{The distribution of pixels in our classification mask for our whole dataset, before data augmentation. The No Fire class contains orders of magnitude more pixels than any other class in the distribution (note the logarithmic scale on the y-axis).}
    \label{fig:class_distribution}
\end{figure}

Due to the significant class imbalance between \textit{No Fire} pixels and those belonging to \textit{Fire}, \textit{Smoldering}, and \textit{Saturated} classes, dataset augmentation was applied to the training set. This augmentation involved horizontally or vertically flipping and adding Gaussian noise exclusively to patches known to contain fire, smoldering, or saturated pixels. By patching the images beforehand, we ensured a more balanced split between \textit{No Fire} and the combined fire-related classes, by selectively augmenting only regions containing fire or smoldering pixels. This approach reduced redundant augmentations of non-fire areas. Figure \ref{fig:class_distribution} depicts the class distribution before augmentation.

% Figure \ref{fig:patch-augmentation} illustrates an example of a patch and the implemented augmentation techniques.

% \subsection{Models}

% Here we describe how we structure our model architectures, and why. This section will be split into two main parts: one describing the classification over the whole patch, and the other running segmentation and FRP prediction on a pixel-level.

\subsection{Classification}
\label{sec:classification}

A crucial initial step is rapidly determining if a patch contains active fire and characterizing its intensity. We formulate this as a four-class classification: \textit{No Fire} (no thermal anomaly), \textit{Smoldering} (moderate thermal signatures, typical of lower-temperature combustion, often found at the edges of the fire or in residual combustion areas), \textit{Flaming} (strong mid-wave infrared anomalies indicative of active burning fronts), and \textit{Saturated} (pixels exceeding sensor limits due to extreme fire intensity or sensor limitations). This coarse classification quickly filters large satellite images, minimizing computational overhead for subsequent, more expensive pixel-level segmentation or FRP prediction.

% \paragraph{Model Architecture and Training} 

% Our classification models are configured for multi-spectral patches of size $24 \times 64 \times 9$, as described in \ref{sec:preprocessing}. 

% However, because our model architectures primarily rely on convolutional operations or self-attention mechanisms that operate on tokenized patch embeddings, these models can be adapted with minimal modifications to different patch sizes or even entire images, allowing for greater flexibility in spatial input dimensions.

First, we employed a custom lightweight 2D CNN optimized for small input dimensions. Additionally, we evaluated deeper and shallower ResNet variants, ultimately deciding on a lightweight ResNet-18, adapted specifically for multi-band data \cite{resnet}. We further explored promising approaches from recent literature, such as MobileNetV3, which employs depthwise separable convolutions and efficient bottleneck structures to factorize standard convolutions into smaller, more efficient matrix multiplications. This factorization substantially reduces computational complexity and memory requirements, making MobileNetV3 particularly suitable for onboard deployment in resource-constrained environments \cite{mobilenetv3}. Finally, we evaluated the Spectral–Spatial Residual Network (SSRN), a specialized architecture that leverages residual connections to effectively capture subtle spectral–spatial relationships inherent in hyperspectral and multispectral imagery, enhancing the model’s capability to discriminate fine-grained features \cite{ssrn}. For MobileNetV3 and SSRN, we utilized the same architectures described in their respective papers (for MobileNetV3 we use the ``MobileNetV3-Small," configuration), and made required modifications for our input layers.

To take advantage of global or long-range context in our data, we further experimented with Vision Transformers (ViT)~\cite{dosovitskiy2020image}. In our ViT implementation, each patch is subdivided into smaller blocks of 6$\times$8 (height $\times$ width) in the spatial dimensions and projected into embedding vectors that incorporate the spatial and spectral dimensions. We embed these vectors with positional encodings so that the transformer’s self-attention mechanism can learn both spatial and spectral relationships. The final output from the transformer encoder is then passed to a classification head.

In preliminary experiments, we also evaluated additional models, including MaskedSST \cite{maskedsst}, Hybrid CNN Transformer, and classical tree-based machine learning baselines (Random Forest \cite{randomforests} and XGBoost \cite{chen2016xgboost}). However, these approaches consistently underperformed compared to the selected models, likely due to small dataset size or the niche datatype of hyperspectral imagery. For conciseness and to maintain focus on the most promising methods with the lowest latency possibility, their detailed results have been excluded from the main discussion.

All of our models were trained using cross-entropy loss, over 30 epochs, a batch size of 128 patches, and with a learning rate of 0.001. These hyperparameters were chosen by performing a grid search, and were found to yield the best results. We used various activation functions across inner layers and architectures, such as Rectified Linear Unit (ReLU) \cite{nair2010rectified}, Leaky ReLU \cite{maas2013rectifier}, Gaussian Error Linear Unit (GELU) \cite{hendrycks2016gaussian}, and H-swish \cite{mobilenetv3}. For all of our models, we used the Adam optimizer \cite{kingma2014adam}. While, due to space constraints, detailed descriptions of the evaluated model architectures are omitted here, we plan to make these details available through an open-source implementation in the near future.

\subsection{Regression and Segmentation}

The regression approach quantifies fire intensity by predicting FRP values for each pixel in multispectral imagery, while the segmentation task assigns a categorical label of \textit{No Fire}, \textit{Smoldering}, \textit{Flaming}, or \textit{Saturated} to each pixel, offering greater nuance than binary detection. After experimenting with various model architectures, we found that a U-Net–style network consistently performed best \cite{unet}. We therefore use U-Net as the underlying model. 

We use these models to compare two inference pipelines: a traditional single-stage pipeline that processes all pixels without filtering, and the PyroFocus pipeline, which first applies a lightweight classifier to identify likely fire-containing patches before forwarding them to specialized U-Net models for detailed segmentation or FRP regression.

% \subsubsection{FRP Regression}
For FRP regression, we employ a Residual U-Net that replaces standard convolution blocks with residual blocks. Each residual block consists of two convolutional layers, batch normalization, ReLU activation, and a skip connection that helps preserve gradients during training. The network’s encoder path uses max pooling to reduce spatial dimensions, and the decoder path upscales via transposed convolutions. We also include deep supervision at multiple scales to encourage better feature learning. A custom loss combines masked mean absolute error (MAE) for focusing on fire pixels, mean squared error (MSE) for handling larger FRP values, and a false positive penalty for discouraging predictions in no-fire areas. 

% This design enables direct regression on an entire patch, while residual connections promote stable training.

% \subsubsection{Segmentation}
For pixel-wise fire segmentation, we adapt the same Residual U-Net backbone, replacing the final regression layer with a multi-class softmax head trained via cross-entropy. The core encoder–decoder structure and skip connections remain unchanged from the FRP version. We trained these models over 30 epochs, with a batch size of 32 patches, inner layer activations of ReLU, and an Adam optimizer with a learning rate of 0.001.

% For FRP regression and image segmentation, we compare the traditional single-stage and PyroFocus pipelines using the same underlying model. The single-stage pipeline performs inference on all pixels without prior filtering. In contrast, PyroFocus first uses a lightweight classifier to identify likely fire-containing patches, forwarding only these to specialized U-Net models for detailed segmentation or FRP regression.

\section{Results}
\label{sec:results}

In this section, we present and discuss the results of our experiments. This will be split up by task, namely the multi-class classification results, the segmentation results, and the FRP regression results.

For sections \ref{results-frp-regression} and \ref{results-frp-segmentation}, inference time comparisons were conducted on the ten images from the test set with the highest fire activity, ensuring that the reported inference times for the second stage reflect the maximum expected runtime across the entire test set. This work was evaluated using a 32GB NVIDIA Tesla V100 GPU.

\subsection{Classification}

\begin{figure}
    \centering
    \includegraphics[width=1\linewidth]{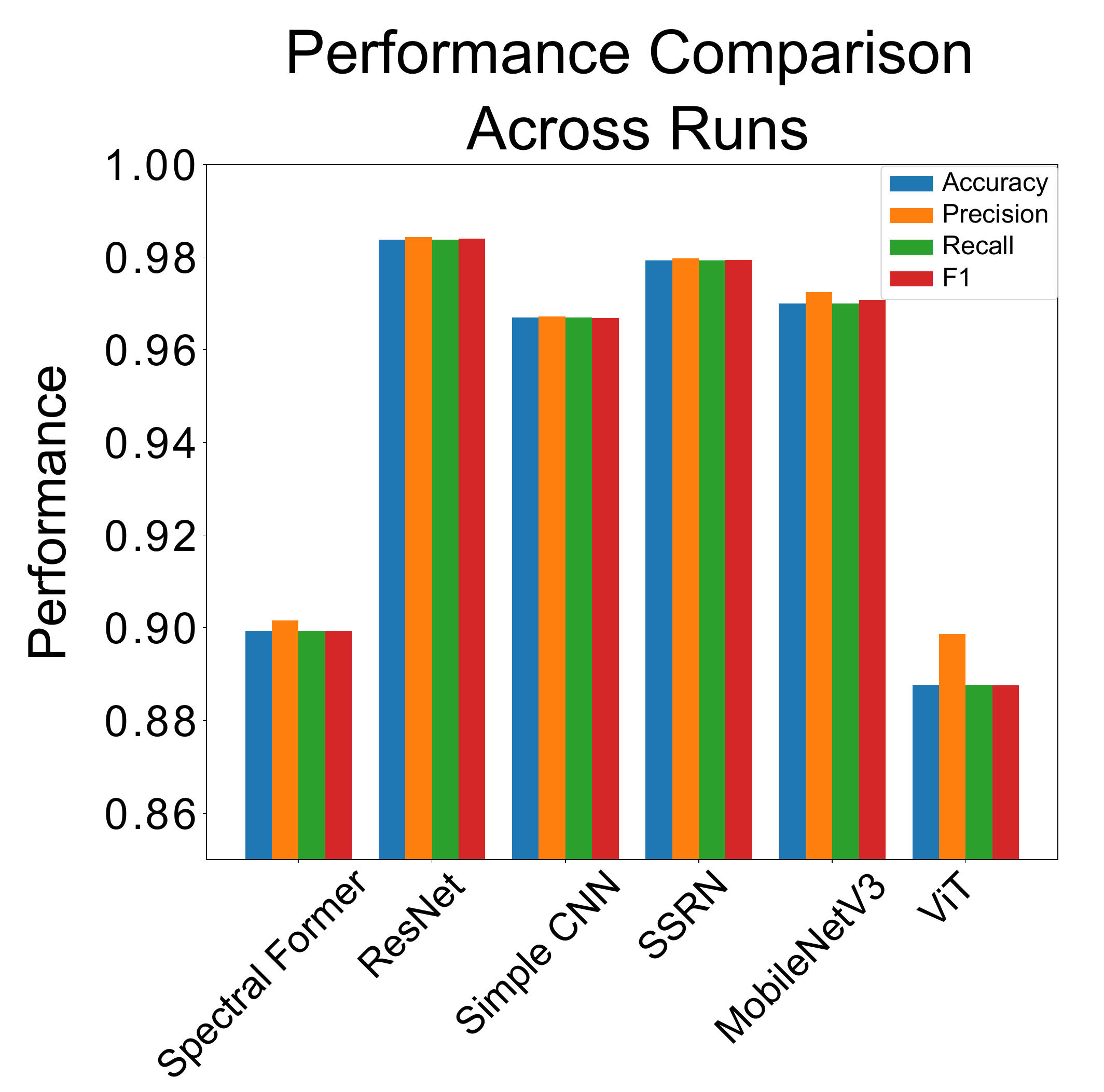}
    \caption{Bar charts comparing accuracy, precision, recall, and F1 score across all models.}
    \label{fig:metrics_bar_charts}
\end{figure}

Figure~\ref{fig:metrics_bar_charts} shows the overall accuracy achieved by each model (Spectral Former, MobileNetV3, Vision Transformer, ResNet, SSRN, and Simple CNN).

% \begin{figure}
%     \centering
%     % \includegraphics[width=1\linewidth]{figures/loss_across_runs.png}
%     \includegraphics[width=1\linewidth]{figures/loss_across_runs.pdf}
%     \caption{Comparison of training, validation, and test loss for each model.}
%     \label{fig:loss_comparison}
% \end{figure}

% Figure \ref{fig:loss_comparison} compares the train, validation, and test losses across the different architectures. 

\begin{figure}
    \centering
    \includegraphics[width=1\linewidth]{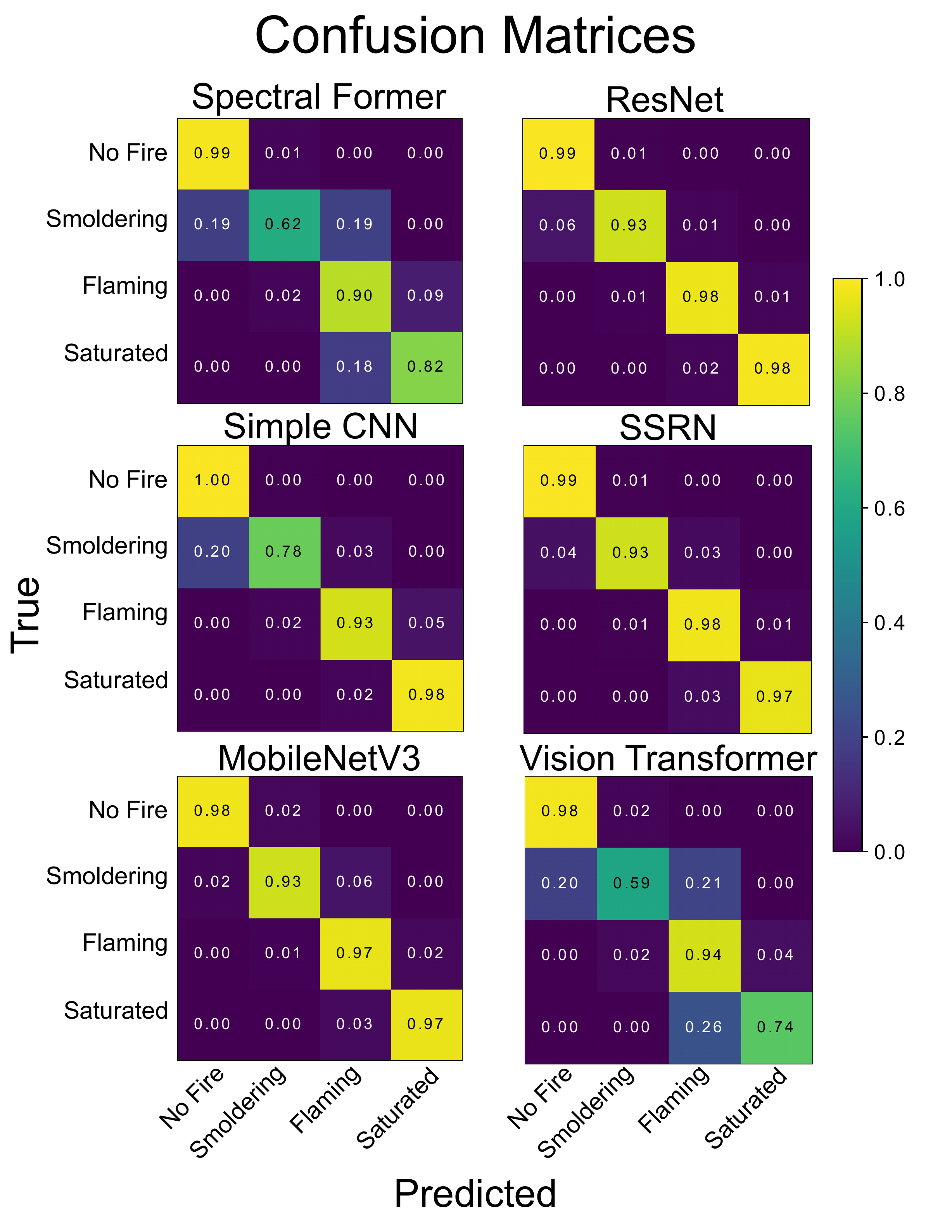}
    \caption{Normalized confusion matrices for each model. 
             Each cell shows the fraction of predictions for a given class (row) 
             classified into each predicted category (column).}
    \label{fig:confusion_matrices}
\end{figure}

The normalized confusion matrices for each model are presented in Figure \ref{fig:confusion_matrices}. Each sub-figure highlights the distribution of predictions across the four classes (\textit{No Fire}, \textit{Smoldering}, \textit{Flaming}, and \textit{Saturated}). The diagonal elements of these matrices illustrate each model’s ability to correctly classify samples of a given class, whereas off-diagonal cells indicate misclassifications.

\begin{table}[ht]
    \centering
    \renewcommand{\arraystretch}{1.2}
    \resizebox{\linewidth}{!}{%
    \begin{tabular}{p{3.8cm} >{\centering\arraybackslash}p{2cm} >{\centering\arraybackslash}p{2.8cm}}
        \hline
        \textbf{Run Name} & \textbf{Inference Time (ms)} & \textbf{Total Parameters (M)} \\
        \hline
        Spectral Former & 3.31 & 21.1 \\
        ResNet & 1.51 & 6.7 \\
        Simple CNN & 0.48 & 1.77 \\
        SSRN & 1.01 & 0.25 \\
        MobileNetV3 & 4.37 & 6.86 \\
        Vision Transformer & 3.16 & 27.7 \\
        \hline
    \end{tabular}
    }
    \caption{Inference time (in milliseconds) and total parameters (in millions) for various models. Tested on a Tesla V100 GPU.}
    \label{tab:model_comparison}
\end{table}

Finally, Table \ref{tab:model_comparison} summarizes each model’s inference time and total parameters.

Considering accuracy, precision, recall, and F1-scores (Figure~\ref{fig:metrics_bar_charts}), confusion matrices (Figure~\ref{fig:confusion_matrices}), and computational efficiency (Table~\ref{tab:model_comparison}), ResNet emerges as the best overall choice, offering an optimal balance between high classification performance across all fire intensity categories and relatively fast inference time. SSRN also demonstrates strong accuracy with even fewer parameters and lower latency, making it highly suitable for extremely resource-constrained scenarios, whereas MobileNetV3, despite good accuracy, incurs significantly higher computational cost. By contrast, Spectral Former and Vision Transformer show inferior intermediate-class performance coupled with higher inference times and parameter counts, limiting their suitability for onboard deployment.

\subsection{FRP Regression}
\label{results-frp-regression}

\begin{figure}
    \centering
    \includegraphics[width=1\linewidth]{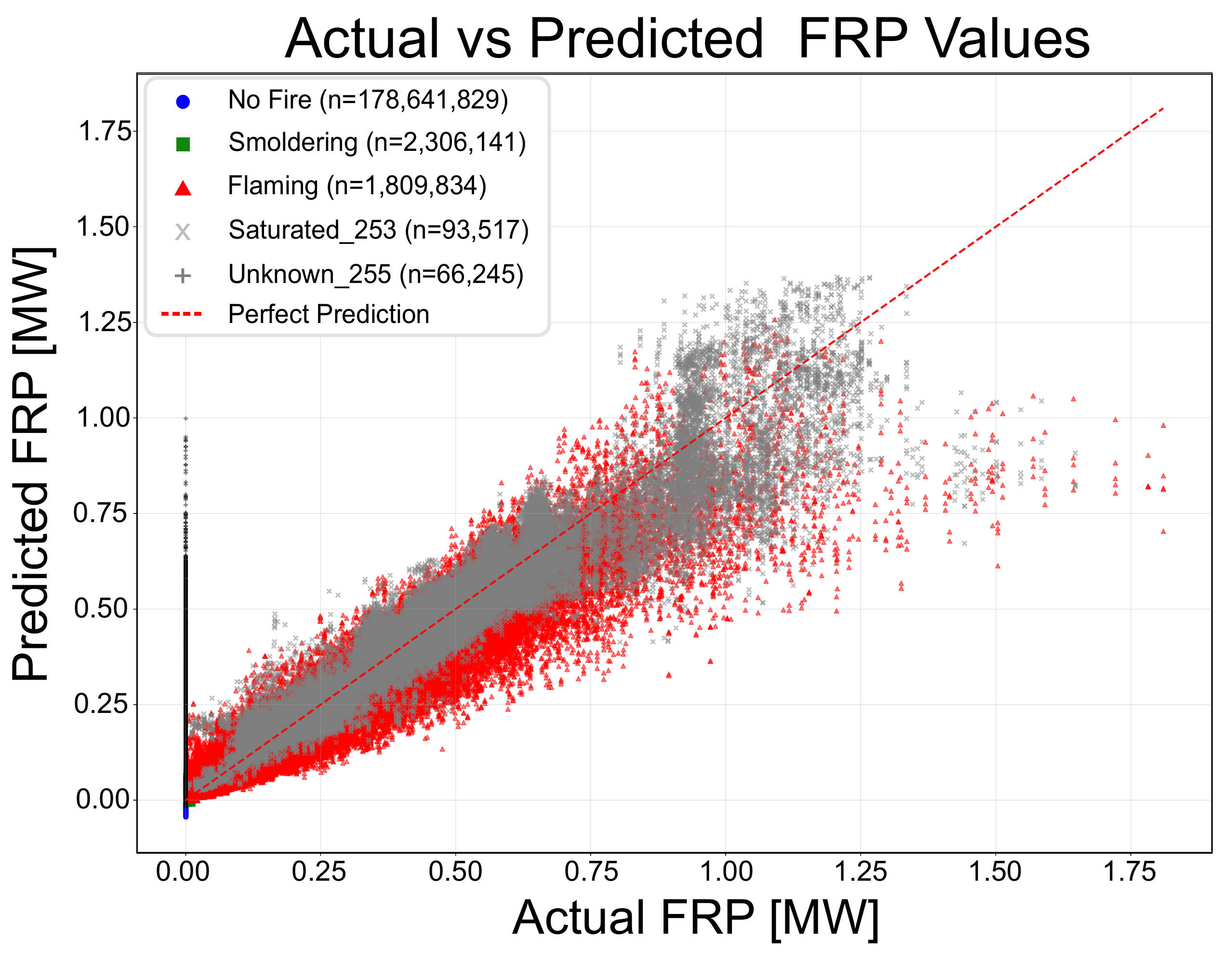}
    \caption{Actual vs. predicted values of pixels for FRP prediction.}
    \label{fig:actual_vs_pred_no_mask}
\end{figure}

Our FRP regression experiments compared the traditional method of a single end-to-end U-Net with our PyroFocus method, which uses two stages to first implicitly mask non-fire pixels, so that the second stage, in this case the U-Net, focuses solely on confirmed fire pixels. As shown in Figure \ref{fig:actual_vs_pred_no_mask}, the traditional end-to-end model generally tracked observed FRP distributions but occasionally produced negative predictions in no-fire areas, suggesting overfitting in regions lacking thermal anomalies.

\begin{figure}
    \centering
    \includegraphics[width=1\linewidth]{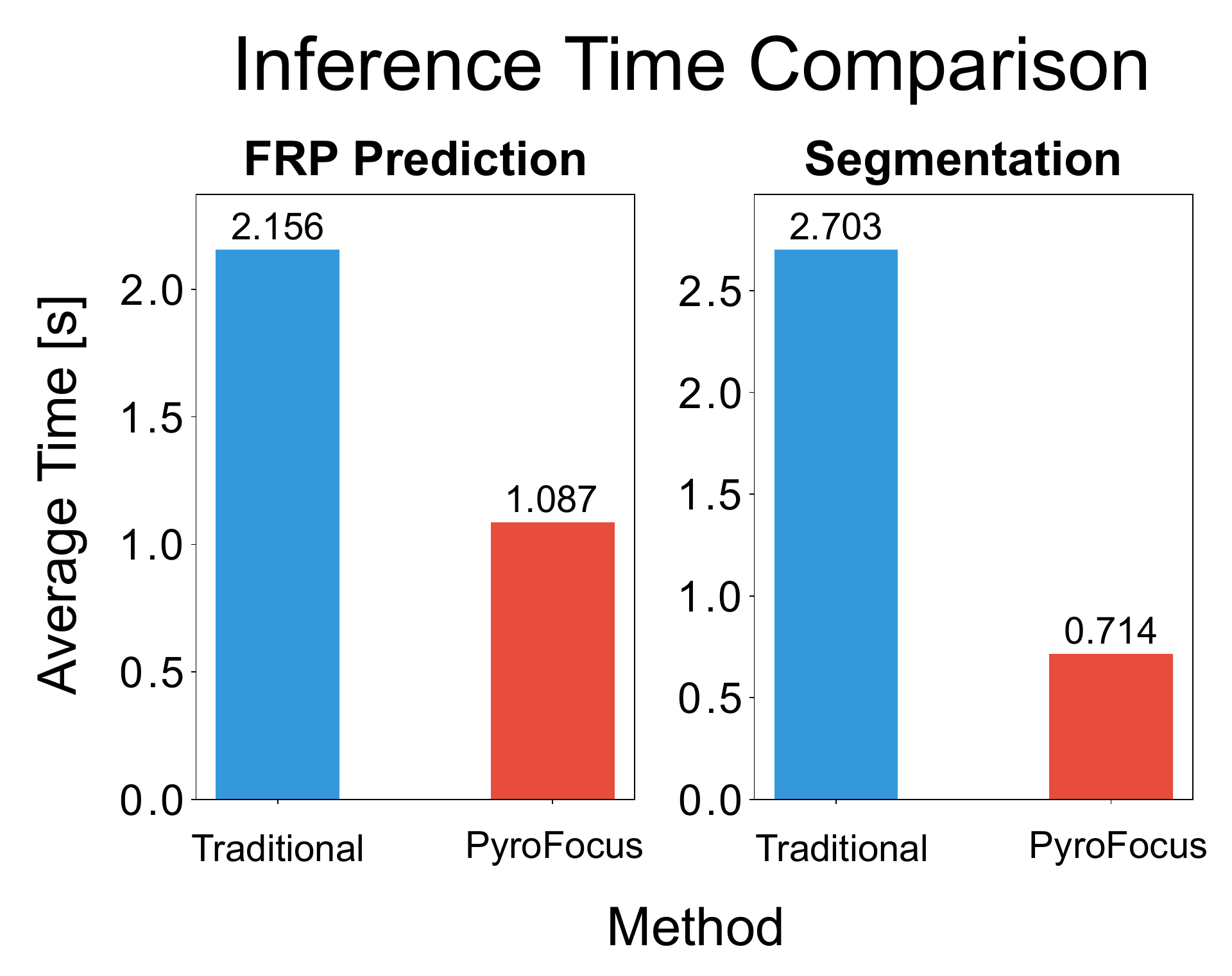}
    \caption{Inference time (in seconds) comparison between the single end-to-end FRP regression and PyroFocus method of FRP regression (left), and the single end-to-end segmentation and PyroFocus method of segmentation (right).}
    \label{fig:inference-time-comparison}
\end{figure}

Figure \ref{fig:inference-time-comparison} shows that the PyroFocus method reduced inference latency by ~1.07 seconds, or a 49.5\% improvement, while maintaining comparable performance metrics (MAE = 0.00349) compared to the baseline.

% With Table
% Our FRP regression experiments compared the traditional method of a single end-to-end U-Net with our PyroFocus method. As shown in Figure \ref{fig:actual_vs_pred_no_mask}, the end-to-end model generally tracked observed FRP distributions but occasionally produced negative predictions in no-fire areas, suggesting overfitting in regions lacking thermal anomalies. 
%Table~\ref{tab:frp_class_results} summarizes overall and per-class FRP metrics, which were similar between the traditional method and PyroFocus. Figure~\ref{fig:inference-time-comparison} shows that PyroFocus reduced inference latency by approximately 1.07 seconds (49.5\% improvement).

%\begin{table}[ht]
%\centering
%\begin{tabular}{lccc}
%\hline
%\textbf{Class} & \textbf{MAE} & \textbf{RMSE} & \textbf{R2} \\
%\hline
%No Fire & 0.0087 & 0.0101 & 0.9857 \\
%Smoldering & 0.0024 & 0.0036 & 0.8489 \\
%Flaming & 0.0111 & 0.0213 & 0.9537 \\
%Saturated & 0.0386 & 0.0574 & 0.9425 \\
%\hline
%\multicolumn{4}{r}{\scriptsize \textbf{Overall:} MAE=0.0086, RMSE=0.0104, R2=0.5912} \\
%\hline
%\end{tabular}
%\caption{FRP regression metrics by class.}
%\label{tab:frp_class_results}
%\end{table}

\begin{figure}
    \centering
    \includegraphics[width=1\linewidth]{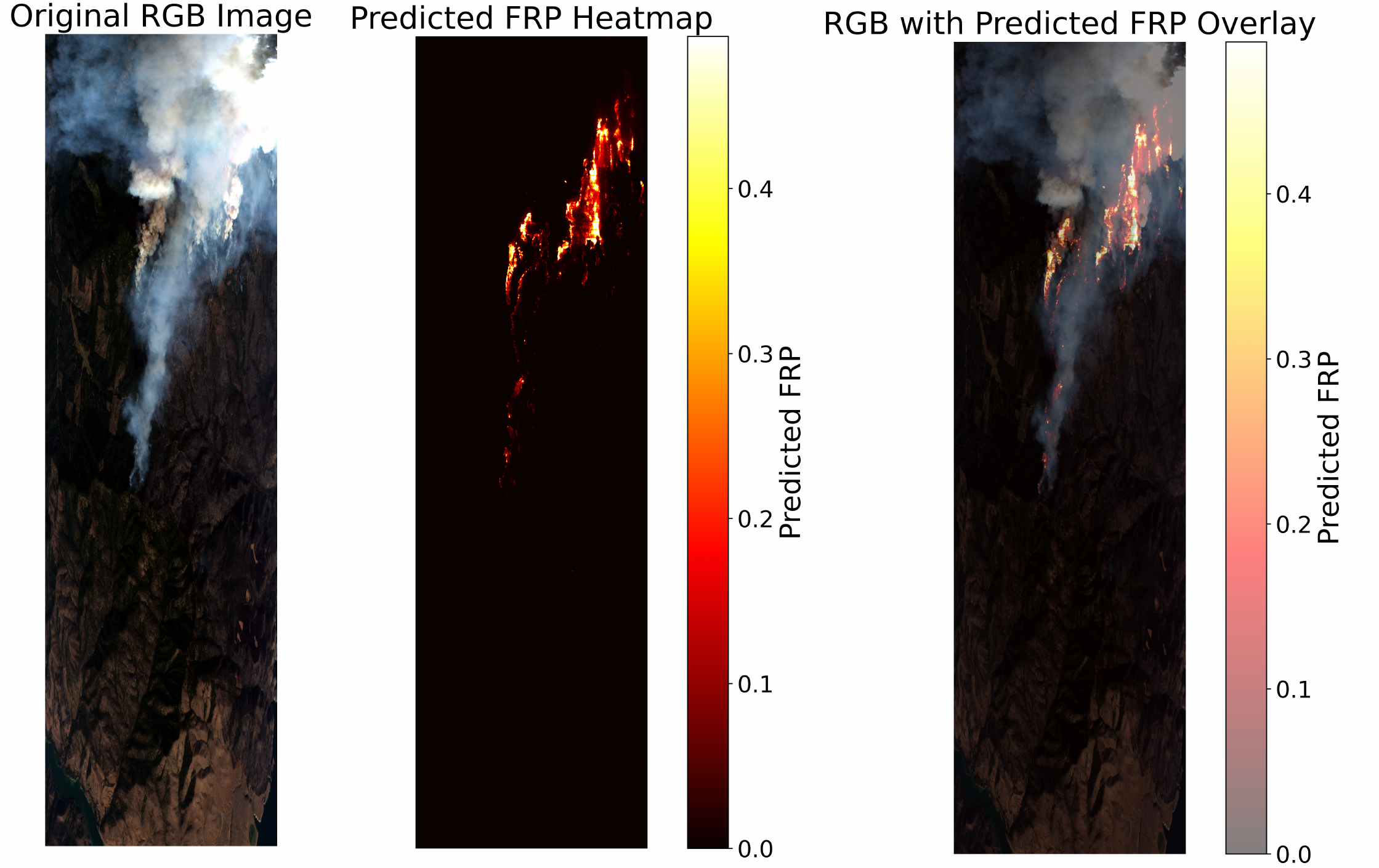}
    \caption{Comparison of an original RGB image of an active wildfire (left), the model’s predicted FRP heatmap (center), and the overlay of the predicted FRP on the RGB image (right). FRP is in MW.}
    \label{fig:twostageregressionoverlay}
\end{figure}

\subsection{Segmentation}
\label{results-frp-segmentation}

\begin{figure}
    \centering
    \includegraphics[width=1\linewidth]{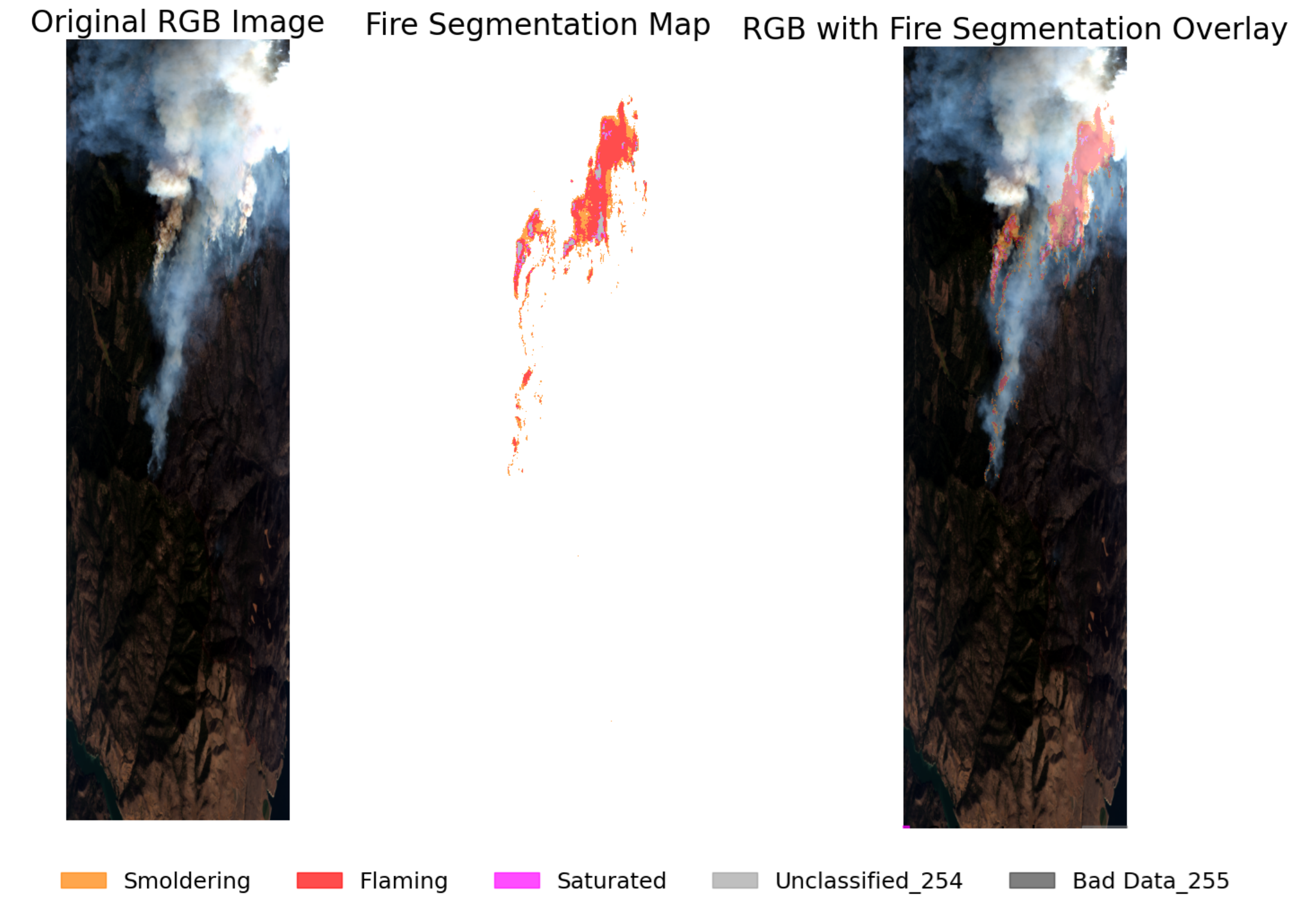}
    \caption{Comparison of an original RGB image of an active wildfire (left), the model’s predicted segmentation mask (center), and the overlay of the segmentation mask on the RGB image (right).}
    \label{fig:twostagesegmentationoverlay}
\end{figure}

Figure \ref{fig:inference-time-comparison} shows that the PyroFocus segmentation approach reduced inference time by an average of 1.989 seconds, or a 73.6\% improvement, compared to the traditional single-stage baseline. In terms of pixel-wise accuracy, both methods correctly identified major clusters of high-intensity pixels (\textit{Flaming} and \textit{Saturated}) and low-intensity regions (\textit{No Fire} and \textit{Smoldering}) at similar rates, but the PyroFocus method processed fewer total pixels and completed segmentation significantly faster. In test samples with partially saturated regions, both models occasionally misassigned some boundary pixels, though overall metrics suggest comparable robustness between the traditional and PyroFocus methods. 

% FireNet \cite{Seydi22} was also trained on the MASTER dataset for a comparison of PyroFocus to the current known state-of-the-art. Table~\ref{tab:FireNetCompare} details the measured inference times (batch size = 64) and mean intersections over unions.

Fire-Net \cite{Seydi22} was also trained on the MASTER dataset to compare PyroFocus with the current state-of-the-art. Table~\ref{tab:FireNetCompare} reports inference time (batch size = 64) and mean IoU.

\begin{table}[h]
    \centering
    \begin{tabular}{c|c|c}
    \hline
        \textbf{Method} & \textbf{MIoU} & \textbf{Inference Time [s]} \\
        \hline
         Fire-Net & 0.9222 & 2.53 \\
         PyroFocus & 0.9916 & 0.71\\
         \hline
    \end{tabular}
    \caption{Inference Times and IoU Model Comparisons.}
    \label{tab:FireNetCompare}
\end{table}

% precision, recall, and F1 values for each class remained comparable between the single-stage and two-stage networks. 
%Table \ref{tab:segmentation-prf-results} summarizes the class-wise metrics. Notably, the test set did not have any samples in the \textit{Saturated} class, so Table \ref{tab:segmentation-prf-results} omits this.

%\begin{table}[ht]
%\centering
%\begin{tabular}{lccc}
%\hline
%%\textbf{Class} & \textbf{P} & \textbf{R} & \textbf{F1} \\
%\hline
%No Fire & 1.0000 & 0.9997 & 0.9998 \\
%Smoldering & 0.9746 & 0.9982 & 0.9863 \\
%Flaming & 0.9980 & 0.9994 & 0.9987 \\
%\hline
%\multicolumn{4}{r}{\textbf{Overall Accuracy: } 0.9996} \\
%\hline
%\end{tabular}
%\caption{Class-wise precision (P), recall (R), and F1-score (F1).}
%\label{tab:segmentation-prf-results}
%\end{table}
\section{Discussion}
\label{sec:discussion}

The primary aim of this work was to explore deep-learning–based approaches for onboard wildfire detection and characterization in resource-constrained environments by leveraging multispectral data from NASA’s MASTER sensor. This section interprets our results and highlights the implications, limitations, and broader insights.

\subsection{Multi-Class Classification}

Our multi-class classification framework targeted four distinct categories (\textit{No Fire}, \textit{Smoldering}, \textit{Flaming}, and \textit{Saturated}) to capture a range of fire intensities without resorting to a simple binary split. Through most of the model architectures tested, several trends emerged that inform how spectral and spatial cues should be leveraged. High overall accuracies masked important nuances related to fire intensity. The confusion matrices in Figure \ref{fig:confusion_matrices} show that most models easily distinguished between no-fire patches and heavily flaming or fully saturated regions, where the thermal signals are unmistakably strong. Smoldering patches posed greater difficulty, often leading to confusion with either low-level flaming or non-fire areas, especially when the visible flaming boundary was not sharply defined. This could be due to subtle spectral–spatial cues that were insufficiently captured by certain architectures.

ResNet and SSRN consistently exhibited the strongest performance across the fire intensity classes, effectively minimizing misclassification errors, particularly for challenging intermediate classes such as \textit{Smoldering} (Figures \ref{fig:metrics_bar_charts} and \ref{fig:confusion_matrices}). SSRN in particular seems quite promising, due to its significantly reduced parameter size (0.25M parameters) and inference time (1.01 ms) (Table \ref{tab:model_comparison}), while also incorporating an attention mechanism and achieving an accuracy of 98.4\%. MobileNetV3 also performed well across these classes, although it showed slightly lower accuracy (97.0\%) compared to ResNet and SSRN (Figure \ref{fig:metrics_bar_charts}). However, the established efficiency and optimization potential of MobileNetV3’s architecture, specifically its use of depthwise separable convolutions and efficient bottleneck structures tailored for resource-constrained edge hardware \cite{mobilenetv3}, remains compelling for practical onboard implementations. Future exploration through quantization, additional architectural tuning, and hardware-specific optimizations may help clarify its true performance. We believe ultimately MobileNetV3 may indeed still be the best model architecture, as the lower performance compared to ResNet and SSRN may be small enough to overcome with further development. Future research, specifically comparing these models after quantization, further architecture design considerations, and hardware-specific optimizations are needed in order to illuminate which model truly has the best onboard performance.

While ViT-based approaches retained global context effectively, Spectral Former and Vision Transformer showed more pronounced difficulties with intermediate classes, particularly with the \textit{Smoldering} class. While Transformer-based architectures could theoretically reduce intermediate-stage misclassifications, our results indicate that smaller or specialized CNNs like ResNet, SSRN, and MobileNetV3 provided a more efficient balance between capturing necessary context and computational performance. This may be due to the nature of Transformer-based architectures requiring significantly more training data to converge properly. Thus, while utilizing more training data can be explored in the future, this can potentially produce a non-favorable trade-off by indirectly increasing the model parameter size to achieve a similar performance, which would ultimately hinder the onboard capabilities of these models.

\subsection{FRP Regression}
\label{sec:regression-discussion}
For FRP prediction, early results suggest that using the PyroFocus method, (i.e., combining classification with a follow-up regression) on only the identified fire pixels can reduce processing overhead without compromising FRP accuracy. Traditional single-stage methods (i.e., regressing on every pixel in an entire scene) unnecessarily devote computational effort to non-fire pixels, most of which remain at or near zero FRP. This leads to higher inference latency, which is an undesirable outcome for airborne or spaceborne missions with real-time constraints.

In addition, because FRP values tend to be heavily skewed as many pixels have relatively low or no fire intensity, while a small fraction saturate the sensor, FRP regression is a challenging learning target. Even slight miscalibrations can lead to large absolute errors for the small subset of very high-intensity pixels. PyroFocus mitigates this imbalance by first determining which areas legitimately need FRP estimation and then applying a regression model calibrated specifically on those fire-containing pixels. Empirically, this approach delivered robust results, most notably in its ability to handle outliers with extremely high FRP levels (Figure \ref{fig:actual_vs_pred_no_mask}). By focusing regression solely on fire-labeled patches, PyroFocus roughly halves the inference time (Figure \ref{fig:inference-time-comparison}) without degrading overall performance. Further evaluation of alternative loss functions and model architectures could potentially yield additional improvements.

% Old table sentence
% As shown in Table \ref{tab:frp_class_results}, the per-class FRP metrics exhibit robustness of the model (e.g., $MAE < 0.012$, R\textsuperscript{2} up to 0.95 for Flaming).

\subsection{Segmentation}
\label{sec:segmentation-discussion}

We observed similar benefits for fire segmentation as in section \ref{sec:regression-discussion} by applying the PyroFocus method. A traditional end-to-end network must process all pixels, including those without thermal anomalies, which increases computation. In contrast, Figure \ref{fig:inference-time-comparison} shows that PyroFocus, which localizes fires before segmentation, significantly reduces inference time by over 70\% while retaining similar segmentation quality (Figure \ref{fig:twostagesegmentationoverlay}). The faster runtime, but similar quality in the PyroFocus approach underscores how a cascade that localizes fires first and then conducts detailed per-pixel labeling can outperform a monolithic model pipeline.  Occasional confusions arose where partial saturation overlapped with truly intense fires in neighboring pixels, but notably, the consistency of \textit{Smoldering} or \textit{Flaming} labels improved when each segmenting model trained on fewer irrelevant pixels. Segmentation maps capture flaming cores and smoldering edges with reduced false positives, underscoring the efficiency of a targeted two-stage pipeline for detailed fire mapping.

% Commented out discussion on table.
% Table \ref{tab:segmentation-prf-results} shows high precision, recall, and F1 ($\ge97\%$), and Table \ref{results-frp-segmentation} indicates an overall accuracy of 0.9996. These results were similar between the traditional single-stage pipeline and PyroFocus. It should be noted, however, that per-class metrics are more meaningful, as even when the first stage filters most patches without fire, there is still an overall class imbalance on the pixel-level.

\subsection{Limitations and Practical Considerations}

Despite encouraging outcomes, there are important limitations to acknowledge. Although we leveraged data from the MASTER 2019 FIREX-AQ campaign, which covers multiple flight lines and fire events, it still represents a finite geographic and temporal scope. Fires in different ecosystems (e.g., boreal versus tropical regions) with varying fuel compositions may exhibit somewhat different spectral responses, limiting generalization. Further work incorporating different datasets is necessary to reduce bias and generalization, and is currently underway by the team.

% \paragraph{Class Overlaps} 
% Our four-class taxonomy (No Fire, Smoldering, Flaming, Saturated) simplifies reality, as real-world fire progressions can exist on a continuum. For instance, edge pixels in a smoldering region may transition intermittently to moderate flaming, complicating the labeling process.

% \paragraph{Sensor Saturation Effects} 
While we identified spectral bands less prone to saturation, extremely intense fires can still produce reflectance values at or near the sensor limit. The \textit{Saturated} class helps capture this phenomenon but does not fully remedy the challenge of losing additional information above saturation thresholds.

Finally, while using the PyroFocus method helps reduce the proportion of obvious non-fire data, during inference there are still abundant background non-fire pixels within patches that do contain fire. This can lead to over-prediction of the non-fire class at the pixel level, artificially inflating performance metrics. Consequently, overfitting on non-fire pixels still remains a concern for the second stage, and further refinement, perhaps through additional training data, will be crucial for robust generalization.

\subsection{Broader Implications} 
The broader significance of these findings extends beyond wildfire detection alone. Our demonstrated pipelines underscore the feasibility of applying advanced deep networks for time-critical remote sensing tasks. As constellations of small satellites and airborne campaigns increase, the need for onboard intelligence that can filter, classify, and prioritize data before downlink is growing. Because our model architectures primarily rely on convolutional operations or self-attention mechanisms that operate on tokenized patch embeddings, these models can be adapted with minimal modifications to different patch sizes or entire images, allowing for greater flexibility in spatial dimensions. Moreover, applying a two-stage detection–regression paradigm could be adapted to other contexts, such as flood monitoring (e.g., identify flood pixels, then regress on water depth) or agricultural assessments (e.g., identify crop stress areas, then quantify damage severity). Our technique of targeted recognition of interesting phenomena can even be extended to the planetary science domain and beyond.

% From a scientific standpoint, the promise of accurate, near–real-time FRP estimates could be transformative for monitoring biomass combustion and its impact on carbon emissions, air quality, and climate forcing. By enabling faster, more granular fire measurements, decision-makers can allocate firefighting resources more optimally, potentially limiting fire spread and mitigating ecological and economic damage.
\section{Future Work}
\label{sec:future_work}

% While our PyroFocus method has shown to save inference time and enhance onboard real-time detection by focusing the most expensive computation on the relatively interesting regions, further work can be done to enhance the performance capabilities. Namely, techniques to improve inference latency and reduce model size while retaining the robust performance. Quantization methods have shown to be successful in drastically reducing model size while retaining similar or same accuracy and other relevant performance metrics \cite{quantizationtransformer, quantizedcnn}. For many hardware platforms, there now exists quantization and other edge computing performance-enhancing techniques should be explored and further benchmarked for this use case. We are exploring quantization, as well as other hardware-specific optimizations, and will be benchmarking our methods on the NVIDIA Jetson Orin NX in the near future.

% From a model architecture standpoint, new model architecture considerations, or varied architecture design and hyperparameter tuning of the same model architectures studied, would be valuable. While the configurations presented in this study were the best we found, the search space remains vast and deep, with new research coming out often that exhibits a new state-of-the-art. We plan to utilize exhaustive model architecture and hyperparameter search techniques, such as the network architecture search (NAS) \cite{mobilenetv3} and bayesian optimization \cite{wu2019hyperparameter}, to assist in finding new state-of-the-art model architectures.

Although our two-stage PyroFocus method can potentially significantly reduce onboard inference time by intelligently focusing computation on relevant regions, further optimizations remain. In particular, quantization and other edge computing techniques have proven effective at shrinking model size while preserving accuracy \cite{quantizationtransformer, quantizedcnn}. We plan to benchmark several of these methods on devices such as the NVIDIA Jetson Orin NX. Beyond hardware-specific improvements, we also see value in refining model architectures. The parameter space of deep networks is vast, and automated searches (e.g., Network Architecture Search (NAS) \cite{mobilenetv3}) and Bayesian optimization \cite{wu2019hyperparameter} can uncover new configurations that outperform our current designs. By combining these hardware and architectural advances, we aim to further enhance PyroFocus for real-time wildfire detection in increasingly resource-constrained environments.
\section{Conclusion}
\label{sec:conclusion}

In this work, we introduced PyroFocus, a two-stage deep learning pipeline designed to perform efficient, real-time wildfire detection and characterization on multispectral data from sensors such as NASA’s MASTER. By separating classification of fire patches from subsequent FRP regression or segmentation, our approach substantially reduces computational overhead without sacrificing accuracy. Through extensive experiments, we demonstrated that this strategy accelerates inference times while retaining accuracy, which can be crucial for onboard inference in airborne and spaceborne platforms.

Moreover, our findings highlight the advantage of leveraging targeted analyses: only regions flagged by the first-stage classifier undergo the more demanding pixel-level prediction, allowing the system to handle larger volumes of imagery under strict resource constraints. By accurately modeling FRP or segmenting fire patches, PyroFocus has the potential to not only improve response times for emergency management, but it also lays the groundwork for robust, fine-grained fire characterization that can inform broader assessments of ecosystem impacts.

Although the approach shows promise across multiple fire regimes, future research should expand the dataset to incorporate geographically and ecologically diverse fire events, further test different model architectures and training techniques, and integrate advanced hardware optimizations (e.g., quantization) to improve model deployment. 

Ultimately, PyroFocus offers a flexible, resource-conscious framework that can be adapted for various real-time remote sensing tasks, thus enabling fast and accurate onboard data processing crucial for more responsive wildfire mitigation efforts and beyond.
{
    \small
    \bibliographystyle{ieeenat_fullname}
    \bibliography{main}
}

\end{document}